\documentclass{article}

\usepackage{biorxiv} 

\usepackage[utf8]{inputenc} 
\usepackage[T1]{fontenc}    
\usepackage{hyperref}       
\usepackage{url}            
\usepackage{booktabs}       
\usepackage{amsfonts}       
\usepackage{nicefrac}       
\usepackage{microtype}      
\usepackage{lipsum}		
\usepackage{lineno}
\usepackage{algorithm}
\usepackage{arevmath}
\usepackage{amsmath}
\usepackage{amsthm}
\usepackage{graphicx}

\usepackage[noend]{algpseudocode}
\usepackage{caption}
\title{Visualizing Data  using GTSNE}

\author{
  Songting Shi
    \\
  Department of Scientific and Engineering Computing\\ 
  School of Mathematical Sciences\\
  Peking University\\
  Beijing 300071, P. R. China \\
  \texttt{songtingstone@gmail.com} \\
}

\begin{document}
\maketitle

\begin{abstract}
We present a new method GTSNE to visualize high-dimensional data points in the two dimensional map. The technique is a variation of  t-SNE that produces 
better visualizations by capturing both the local neighborhood structure and the macro structure in the data. This is particularly important for high-dimensional data that lie on continuous low-dimensional manifolds. We illustrate the performance of GTSNE on a wide variety of datasets and compare it the state of art methods, including t-SNE and UMAP. The visualizations produced by GTSNE are  better than those produced by the other techniques on almost all of the datasets on the macro structure preservation.
\end{abstract}


\section{Introduction}
High-dimensional data visualization is a very important problem for human to sense the data. Currently, the state of art methods are t-SNE (\cite{t-SNE}, \cite{BH-SNE}) and UMAP (\cite{UMAP}), which has similar principle for the nonlinear low dimension reduction. They use neighborhood  probability distribution to connect the high-dimensional data points to low-dimensional  map points, which try to make the local relative neighborhood relation unchanged but ignoring the change in the macro structure of the data. 
However, this may make the low dimension map points  representing the high-dimensional structure unfaithfully. In the low-dimensional neighborhood keeping and patching process,  t-SNE sometimes will make the neighborhood relations in the high-dimensional structure break in the  the low-dimensional space.
We add a macro loss term on the loss of t-SNE to make it keep  the relative  k-means centroids structure   in the low and high dimensional space, which basically keep the macro structure unchanged in the low dimensional space.

\section{Methods}
We now begin to derive the loss function of the global t-distributed stochastic  neighborhood embedding (GTSNE). Suppose that there are $N$ points in the high-dimensional space, $X = \{ x_1, x_2, \ldots, x_N\}$, where $x_i \in \mathbb R^D, \; i = 1, 2, \ldots, N$.
We want to get their low-dimensional embedding map points, $Y = \{ y_1, y_2, \ldots, y_N \}$, where $y_i \in \mathbb R^d$, where $d=2$. 

\subsection{The Loss of t-SNE}
Recall that the loss function of  t-SNE  is given by 
\begin{equation}
L_{t-SNE}(Y \;| \; X) = KL(P||Q)  = \sum_{i,j=1, \; i\ne j}^N p_{ij} \log \frac{p_{ij}} {q_{ij}}
\end{equation}
where $p_{ij}$ is the probability of high dimensional point $x_i$ connecting to  $x_j$, and $q_{ij}$ is the probability of low dimensional point $y_i$ connecting to $y_j$. The probability $p_{ij}$ and $q_{ij}$ characterize the neighborhood relation of $i$ and $j$. The close two points will have a higher probability than those two far-separated points. The key is that we need to seek the probability distributions in both the high-dimensional space and low dimensional space, such that they can match each other, i.e. when $p_{ij} = q_{ij}$ we will have the best layouts $Y$  in the low dimensional space.  

t-SNE use the Gaussian probability to model the neighborhood relations in the high dimensional space, i.e.
\begin{equation}
\tilde p_{ij} = \left \{
\begin{array}{cl}
 \frac{1} {Z_{x,i}} \exp( - \frac {||x_i - x_j||^2}{2\sigma_i^2}) & \text{if}   j \in i's \text{ K-nearest neighbors} \\
 0 & \text{otherwise}
 \end{array}
 \right .
\end{equation}
\[Z_{x,i} =  \sum_{j \in i's \text{ K-nearest neighbors}} \exp( - \frac {||x_i - x_j||^2}{2\sigma_i^2}) \].
\begin{equation}
\label{eq:P}
 \begin{array}{c}
p_{ij} = \frac{\tilde p_{ij} + \tilde q_{ij} } {2N} 
\end{array}
\end{equation}
where the $\sigma_i$ was chosen such that it satisfies the perplexity equation
\begin{equation}
\begin{array}{c}
2^{ H(\tilde p_{ij}) }= Perplexity \\
H(\tilde p_{ij}) = - \sum_{j \in i's \text{ K-nearest neighbors}} \tilde p_{ij} \log (\tilde p_{ij})
\end{array}
\end{equation}
. Intuitively, this means that the probability $\tilde p_{ij}$ can effectively distinguish $\text{Perplexity}$ neighbors of $i$. 

To solve the crowding problem, i.e. when make that the distance relation keeping in the low dimensional space, it will make that the mediately separate  points in the high dimensional space  clustering together in the low dimension space, t-SNE uses the heavy tail  t-Distribution to model the low dimensional neighborhood relations. 
\begin{equation}
\begin{array}{c}
\tilde q_{ij} = \frac{1} { 1 + ||y_i - y_j||^2}\\
Z_{y}=  \sum_{i \ne j }  \tilde q_{ij} \\ 
q_{ij} = \frac{\tilde q_{ij}} { Z_{y }} 
\end{array}
\end{equation} 
for $i \ne j$, and $q_{ii}  = 0$. 

In the above formulation, t-SNE only captures the local neighborhood relations in the low dimension embeddings. In our numerical experiments, we find the t-SNE map points can not faithfully represent the high-dimensional data points. 
There exist two problems. 
The first one is that 
t-SNE can not fully preserve the local neighborhood relation.  
This occurs when two neighbor points were separated by a line in the map points in the low dimension layout, t-SNE will separate the two points on the two sides of the line and push them far away from the line. 
Note that the problem is  due to that the t-SNE loss is non-convex, which is hard to optimize. Once a line lies in the middle of the near points, it is hard to push the line far away from the two points. 
The second problem is that t-SNE can not preserve the macro structure of data, e.g, it will project a three dimensional sphere into 2D space but do not own a circle boundary. 
To overcome the above two problems, we propose the following GTSNE loss, which will consider both the local neighborhood structure, and also the macro structure of the data points. 

\subsection{Global t-Distributed Stochastic Neighbor Embedding}
To characterize the global structure of the high dimensional points, we use the k-means clustering centroids in the high dimension space, their neighborhood relations and the data point with the centroids relations to represent the macro structures.
To do this, we first run PCA on $X$ to get the PCA embedding $Z = \{z_ 1, z_2, \ldots, z_N \}$, where $z_i \in \mathbb R^{D_Z}$.
Then we run k-means clustering algorithm on $ Z$ to get the k-means centroids $  T = \{ t_1, t_2, \ldots,  t_K\}$, where $t_i \in \mathbb R^{D_Z}$. 
The k-means centroids capture the global structure of the data points. 
For each point $i$, we calculate the probability that point $i$ belong to  the cluster $k$ by the t-Distributed distribution denoted by $R_{k,i}$, 
\begin{equation}
\label{eq:R}
\begin{array}{c}
\tilde R_{ki} = \frac{1} {1 +\frac{d^2}{D_Z^2} ||z_i - t_k||^2}\\
Z_{z, i }=  \sum_{k}  \tilde R_{ki} \\ 
R_{ki} = \frac{\tilde R_{ij}} { Z_{z, i }} 
\end{array}
\end{equation} 
Note that we use the scaling factor $\frac{d^2}{D_Z^2}$ on the distance $||z_i - t_k||^2$, since this $R_{ki}$ will use to represent the data point $y_i$ belong to the its cluster centroids $c_k$ in the low dimensional space.

To transfer the global structure information in the the low dimension map points, we use the t-distributed distribution to characterize these $K$ centroids relations, 
\begin{equation}
\label{eq:P-macro}
\begin{array}{c}
\tilde p_{macro, kl} = \frac{1} { 1 +  ||t_k - t_l||^2}\\
Z_{t }=  \sum_{k,l=1,\; k\ne l}^K  \tilde p_{macro, kl} \\ 
p_{macro, kl} = \frac{\tilde p_{macro, kl} } { Z_{t }} 
\end{array}
\end{equation} 

To characterize the low dimensional macro structure, we define the low dimension centroids by $R$ with the formula,
\begin{equation}
\begin{array}{c}
c_{k} = \frac { \sum_i R_{ki} y_i } {\sum_i R_{ki}}, \; k=1,2, \ldots, K
\end{array}
\end{equation} 
And define the corresponding low-dimensional t-Distributed macro neighborhood relations by 
\begin{equation}
\begin{array}{c}
\tilde q_{macro, kl} = \frac{1} { 1 + ||c_k - c_l||^2}\\
Z_{c }=  \sum_{k,l=1,\; k\ne l}^K  \tilde q_{macro, kl} \\ 
q_{macro, kl} = \frac{\tilde q_{macro, kl} } { Z_{c}} 
\end{array}
\end{equation} 
for $k \ne l$ and $ q_{macro, kk} =0$.

Now we can get the GTSNE loss function, 
\begin{equation}
\begin{array}{ll}
L(Y)&= L_{micro} + \alpha L_{macro} + \beta L_{k-means}\\
  &= KL(P||Q) + \alpha KL(P_{macro} || Q_{macro}) + \beta L_{k-means}\\
  &= \sum_{i,j=1,\; i \ne j}^N p_{ij} \log \frac{p_{ij}} {q_{ij}}  \\
  & \quad + \alpha  \sum_{k,l=1,\; k \ne l}^K p_{macro, kl} \log \frac{p_{macro, kl}} {q_{macro, kl}}   + \beta \frac {1} N  \sum_{k,i} R_{ki} || y_i - c_k||^2
\end{array}
\end{equation} 
where $\alpha, \beta \in \mathbb R$ are weight parameters of the loss.  The loss was composed of three parts. The first part is the t-SNE loss $L_{micro} = KL(P || Q)$ which penalizes the mismatch between $P$ and $Q$, such that $Q$ will maintain the local neighborhood relation. The second part is the macro loss $L_{macro} = KL(P_{macro} || Q_{macro})$, which try to make the low dimensional centroids relations match the high dimensional centroids relations. The third part is the k-means loss $L_{k-means}$, which try to make that the map points $y_i$ satisfies the centroids belong relations $R_{ki}$.

After some mathematical calculation, we get the gradient of loss $L(Y)$, 
\begin{equation}
\label{eq:gradient}
\begin{array}{ll}
\frac { \partial L( Y) } {\partial y_i}&= 4 \sum_{j\;: \;j \ne i} (p_{ij} - q_{ij})\tilde q_{ij} (y_i - y_j) \\ 
& \quad + 2\alpha \sum_{k,l;\; k\ne l} (p_{macro, kl} - q_{macro, kl} ) \tilde q_{macro, kl} (R_{ki} - R_{li})(c_k - c_l) \\
& \quad + 2 \beta \frac 1 N  \sum_k R_{ki}(y_i - c_k)
\end{array}
\end{equation} 

We use the gradient descent method to optimize the loss function. The adaptive learning rate scheme described by Jacobs \cite{Jacobs1988}  was used, which gradually increases the learning rate in the direction in which the gradient is stable.

Now we give the GTSNE algorithm~(\ref{alg:gtsne})  to guide the details of imagination. 
\begin{algorithm}
\caption{GTSNE: Global t-distributed Stochastic Neighbor Embedding}
\label{alg:gtsne}
\begin{algorithmic}[1]

\Function{GTSNE}{$X$, perlexity, $N$,  $K$, $D$, $D_Z$, $d$} 
   \State \textbf{Dataset} $X = \{ x_1, x_2, \ldots, x_n \} $. 
   \State cost function parameters: perplexity $Perp$, weight parameter $\alpha$ of the macro loss $L_{macro}$, weight parameter $\beta$ of the k-means loss. 
   \State learning rate $\eta$, the momentum scalar $\gamma$. 
   \State \textbf{Result}: low-dimensional data representation $ Y = \{y_1, y_2, \ldots,  y_n \}$.

   \State Sample initial solution  $ Y = \{y_1, y_2, \ldots,  y_n \}$ from   $\mathcal N (0, 10^{-4}I) $
   \State Initializing the moment accumulate gradient $u_Y \in \mathbb R^{N \times d}$ with values $0$. 
    \State Compute $\Sigma = X^T X$, do the SVD decomposition $\Sigma = U\Lambda U^T$. Get the PCA embedding $Z = XU[:,0:D_Z]$. Run k-means algorithm on $Z$ with number of clusters $K$. Get the k-means centroids $T \in \mathbb R^{C \times D_Z}$.Compute the cluster assignment probability matrix $R$ by equation (\ref{eq:R}). Compute the macro probability matrix $P_{macro}$ by the equation (\ref{eq:P-macro}) 
    \State Search the K nearest neighbors for each $x_i$ with Euclidean distance $d_{ij} := || x_i - x_j||^2$ which finished by the vantage point tree algorithm(\cite{vptree}). 
    \State  Compute the the probability matrix $P$ by equation (\ref{eq:P}).

	   \Repeat
            \State Compute the gradient $g_Y \in \mathbb R^{N \times d}$ of $Y$ with $g_{y_i }=\frac { \partial L( Y) } {\partial y_i} $ as given in equation (\ref{eq:gradient})
            \State Update the gains of gradient with $\text{gains}_{y_i}  = (sign(g_{y_i}) != sign(u_{y_i}) ? (\text{gains}_{y_i} + 0.2) : (\text{gains}_{y_i} * 0.8), \; i=1, \ldots, N$.
            \State Update the momentum accumulated gradient $u_{y_i} =  \gamma * u_{y_i} - \eta * \text{gains}_{y_i} *  g_{y_i}, \; i= 1, \ldots, N$. 
            \State Update $Y$ with $y_i = y_i + u_{y_i}, \; i=1, \ldots, N$.
   \Until{convergence}
            	
   \State \Return{$Y$}
\EndFunction

\end{algorithmic}
\end{algorithm}

 \textit{ \textbf {Implementation details.} } We use the quadratic tree (\cite{BH-SNE})  to compute the t-SNE gradient part $\sum_{j\;: \;j \ne i} (p_{ij} - q_{ij})\tilde q_{ij} (y_i - y_j)$  approximately. 

\section{Experiments}
To compare performance of GTSNE , we compare it with PCA, t-SNE and UMAP algorithms,  on  both the  simulation data and real data. 

The parameters are set to the default value for each algorithm. 
\begin{itemize}
\item PCA from sklearn.decomposition.PCA.
\item GTSNE: $\alpha = 10^{-2}$,  $\beta = 5*10^{-2}$, $Perp=30$, $K=90$.
\item t-SNE from sklearn.manifold.TSNE: $Perp=30$. 
\item UMAP from umap.UMAP: "n\_neighbors"= 30, "min\_dist"= 0.3.
\end{itemize}

\subsection{Simulation Data}
\begin{figure}
  \centering
\includegraphics[width=150mm]{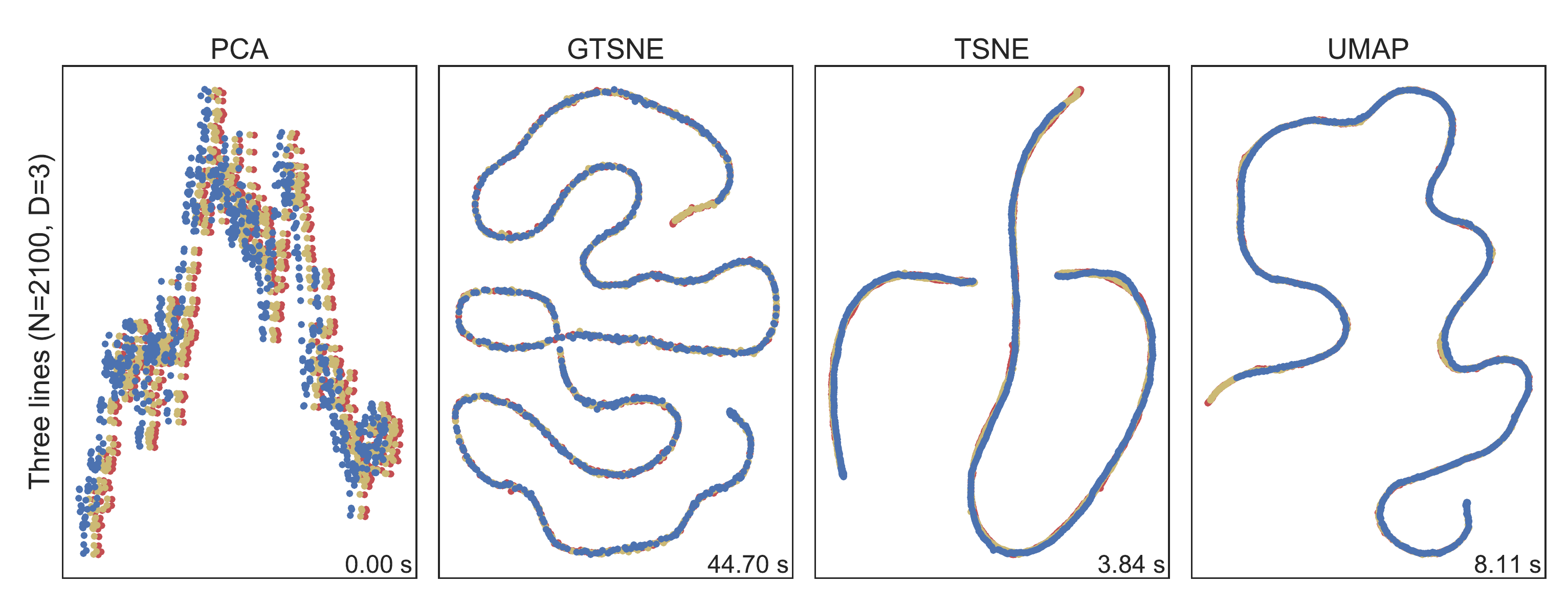}
  \caption{There lines dataset, with dimension $N=2100, D=3$.}
\label{fig:simulation}
  \end{figure}

We first run the algorithm on the simulated data to verify the effectiveness of GTSNE. The simulated data are three continuous lines in the high dimension data, which was generated by 
\begin{enumerate}
\item  Generate the  velocity $V \in \mathbb R^{N_s \times D}$ by random sampling from the normal distribution, i,e, $ V_{il} \sim \mathcal N(0,36),  \; i=1 \ldots, N_s; \; l = 1, \ldots, D$.
\item  Choose three start points of data points $x_{start,1} = \mathbf 0$, $x_{start,2}  =  50* \mathbf 1$.
    $x_{start,3}  =160 * \mathbf 1$.  where $ \mathbf 0$ is the zeros vector with length $D$  and $\mathbf 1$ is the ones vector with length $D$. 
\item  Generate the data points $X \in \mathbb R^{N \times D}$  where  $N = 3N_s$ from the three starting points and moving along the velocity $v_i$ one by one. i.e. 
\begin{equation*}
\begin{array}{l}
    	x_1 = x_{start, 1}\\
	x_{N_s+1} = x_{start, 2}\\
	x_{2N_s+1} = x_{start, 3}\\
	x_{i+1} = x_i + v_i, \; i = 1, \ldots, N_s-1 \\
	x_{N_s + i+1} = x_{N_s + i}  + v_{i}, \; i = 1, \ldots, N_s-1 \\
	x_{2N_s + i+1} = x_{2N_s + i}  + v_i, \; i = 1, \ldots, N_s-1   	
\end{array}
\end{equation*}
\end{enumerate}
We take $N=2100$ and  $D=3$ to generated  the data. After running the dimension reduction methods, we get the  results showed in Fig~\ref{fig:simulation}. The figure shows that t-SNE break the lines while GTSNE and UMAP do keep the lines continuity, which shows the effectiveness of the GTSNE. 

Why t-SNE break the continuous line? From the figure, we see that two horizontal  neighbor points are separated by the vertical line. After t-SNE run into this state, the gradient of t-SNE at one of the neighbor points was driven by two forces. The attractive force comes from their neighbor points, 
 which will make this point close to them. The another repulse force
  comes the points on the the middle  lines, which will make that the point far from them. 
  When the two forces balanced with each other, i.e. canceled to zero. The point do not move when the algorithms running. Thus t-SNE will jump into the local minimum of loss, and can not jump out  from it by the gradient descent. 
 When in the loss of GTSNE, the macro loss part will strength the attractive force of two neighbor points, since if the do not close to each other, the centroid probability will do not match to there high dimensional parts, so that it will make the continuity of the lines.

But in the figure, we also see that GTSNE twists the lines in the low-dimension map. This need to be improved, which is left to the future work. 

\subsection{Real Data}

\subsubsection{Five toy datasets}
Now we run the algorithm on five famous toy datasets. Their information are summarized in Table~{\ref{tab:toy}

\begin{table}
 \caption{The $5$ toy example,  where $C$ the number of classes if it has.}
  \centering
   {
  \begin{tabular}{lll}
    \toprule
    Dataset & Dimension      & Description \\
    \midrule
     Blobs & $N=500, \; D=10, \; C= 5 $ &Isotropic Gaussian blobs 	\\
     Iris & $N=150, \; D=4, \; C= 3 $ & The iris dataset 	\\
     Wine & $N=178, \; D=13, \; C= 5 $ &The wine dataset \\ 
     Swiss Roll & $N=1000, \; D=3 $ &Swiss roll dataset \\ 
     Sphere & $N=600, \; D=3 $ &The sphere dataset \\ 
    \bottomrule
  \end{tabular}
  }
  \label{tab:toy}
\end{table}
After running the algorithms, we get the results showed in Fig~\ref{fig:toy-example}.

  \begin{figure}
  \centering
\includegraphics[width=150mm]{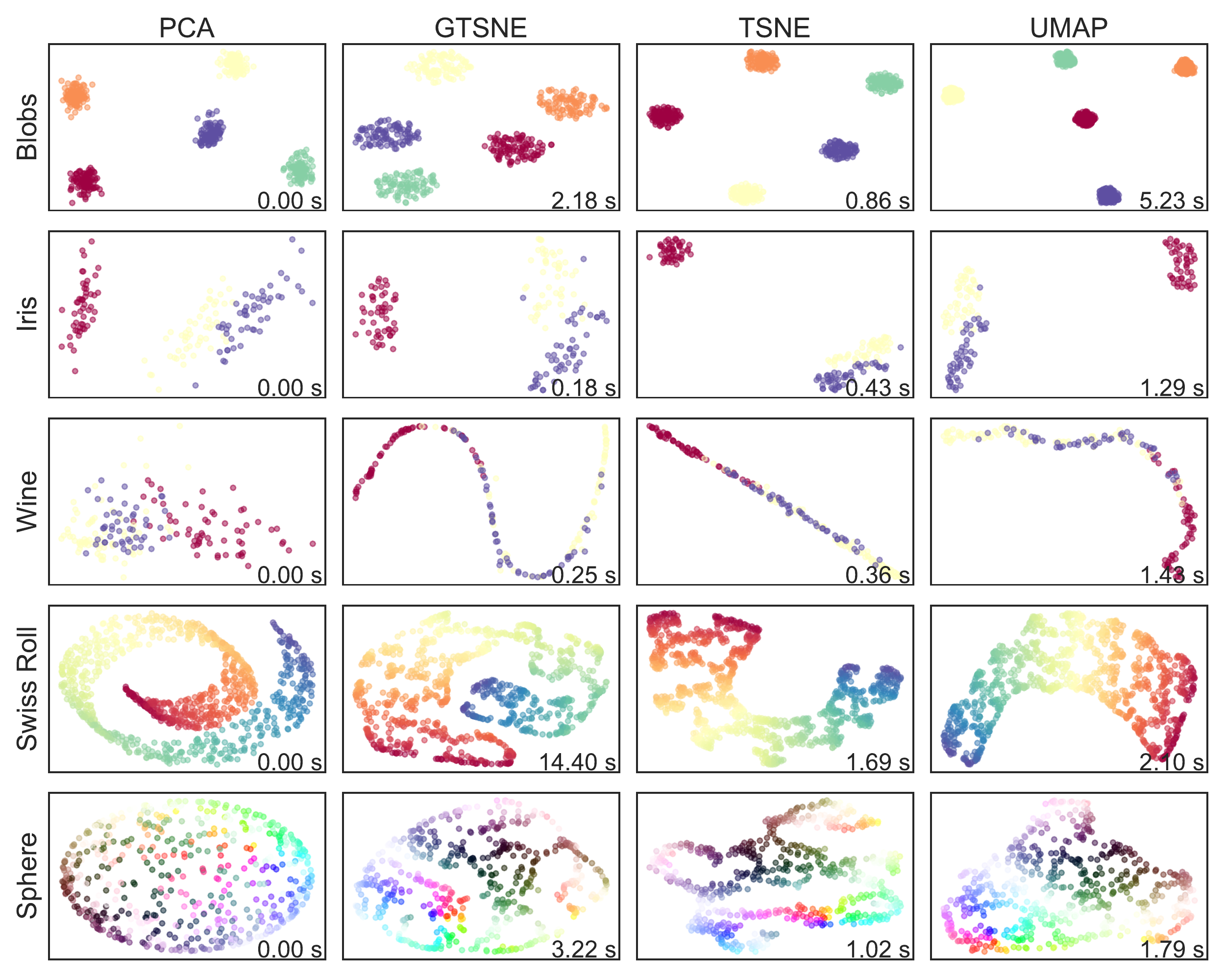}
  \caption{Five toy datasets.}
  \label{fig:toy-example}
  \end{figure}
  
From the results we see that GTSNE worked well on these datasets. On the Swiss Roll dataset,  GTSNE preserves the continuous circle structure while t-SNE and UMAP only get the half circle. On the Sphere dataset,  GTSNE  preserves the sphere shape which are better than the results of  t-SNE and UMAP.

\subsubsection{MNIST dataset}
The MNIST database of handwritten digits $0, 1,\ldots, 9$, has a training set of 60,000 examples, and a test set of 10,000 examples. Each example is an image of $28 \times 28$ pixels. The whole dataset contains $70,000$ examples. 
%
%
 
 \begin{figure}
  \centering
\includegraphics[width=150mm]{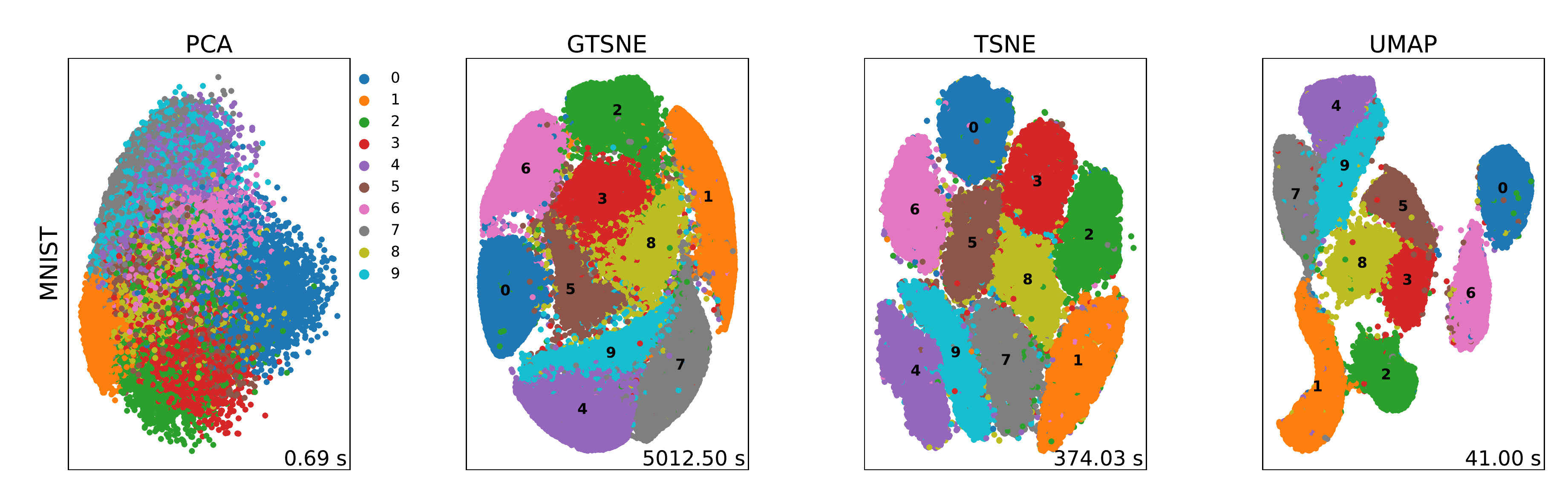}
  \caption{MNIST dataset with dimensions $N=70,000, \; D=784$}
  \label{fig:mnist}
  \end{figure}
  From the result, we see that GTSNE do a comparative representation with t-SNE and UMAP.
%
\subsubsection{Pancreas dataset}
We now run the algorithms on the Pancreas dataset, this dataset was used in \cite{dynamical-RNA-velocity}. It is a single cell RNA-seq dataset.  After selecting the velocity genes, we get the final dataset which has the dimension $3696 \times 2000$, i.e. $3696$ cells and $2000$ selected velocity genes. After running the algorithms, we get the results showed in Fig~\ref{fig:pancreas}. 

    \begin{figure}
  \centering
\includegraphics[width=150mm]{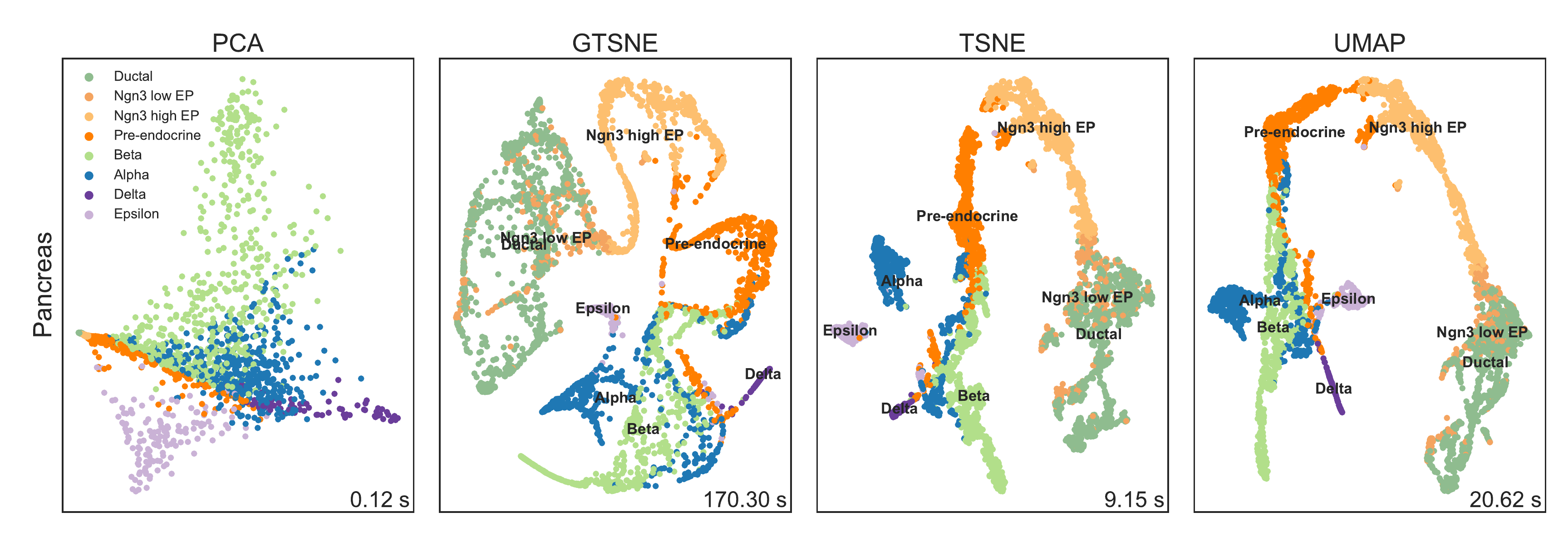}
  \caption{The Pancreas dataset, with dimension $N=3696, D=2000$.}
  \label{fig:pancreas}
  \end{figure}
  
  From the results, we see that GTSNE works similarly with t-SNE and UMAP.  And GTSNE  generates a continuous map which is similar to UMAP, while the result of t-SNE has some breaks in the continuous structure.

\section{Discussion}
GTSNE use the k-means method on the PCA embedding of the high dimensional data to grasp the macro structure, and try to preserve the relative relations of centroids by probability in the low dimensional space. But it has some limitations.

%
 The first problem of GTSNE is that is run slowly in the large dataset, for the MNIST dataset ($N=70,000, D=784$), it takes about one and half hour. It need to make more efficient implementation. 
  
The second  essential question is that  how to define the macro structure? In this paper, we use the k-means centroids to characterize the macro structure, it is just an initial try. Can we find more reasonable solutions? The answer need to find by the reader. 

\section{Conclusion}
In this paper, we proposed a visualization method --- GTSNE, which  is a modified version of t-SNE.  It include the macro structure in the loss function to make that the low dimensional map preserve the macro structure. We hope that this method will help to the data visualization which need to preserve the macro structures.

\section*{Acknowledgements}
Thank to my family ( especially for my mother, Qixia Chen and father, Wenjiang Shi )  for they provides me a suitable environment for this work. Thank to all the teachers who taught me to guide  me to the road of truth.

\section*{Appendix A. Code availability }
GTSNE are available as python package on~\url{https://github.com/songtingstone/gtsne}. 
Scripts to reproduce results of the primary analyses will be made available on~ \url{https://github.com/songtingstone/gtsne2021}.
The code is learned and adapted the C implementation~\url{https://github.com/danielfrg/tsne}  of  BH-SNE (\cite{BH-SNE}), special thanks to Laurens van der Maaten and Daniel Rodriguez. 

\section*{Appendix B. Derivation of the GTSNE gradient. } 
The derivation of GTSNE gradient is similar to the derivation of t-SNE gradient. We now give the details of the derivation. 
The loss function of GTSNE consists of three parts,
\begin{equation*}
\begin{array}{ll}
L(Y)&= L_{micro} + \alpha L_{macro} + \beta L_{k-means}\\
  &= KL(P||Q) + \alpha KL(P_{macro} || Q_{macro}) + \beta L_{k-means}\\
  &= \sum_{i,j=1,\; i \ne j}^N p_{ij} \log \frac{p_{ij}} {q_{ij}}  \\
  & \quad + \alpha  \sum_{k,l=1,\; k \ne l}^K p_{macro, kl} \log \frac{p_{macro, kl}} {q_{macro, kl}}   + \beta  \frac 1 N \sum_{k,i} R_{ki} || y_i - c_k||^2
\end{array}
\end{equation*} 

 so does its gradient,
\begin{equation}
\label{eq:grad-total}
\begin{array}{ll}
\frac { \partial L(Y) } {\partial y_i} &=\frac { \partial L_{micro} } {\partial y_i} +   \alpha \frac { \partial L_{macro} } {\partial y_i}  + \beta \frac { \partial L_{k-means} } {\partial y_i} 
\end{array}
\end{equation} 

The first part is
\begin{equation}
\begin{array}{ll}
\frac { \partial L_{micro} } {\partial y_i}  &= \frac { \partial KL(P||Q)  } {\partial y_i}\\
&=  \frac { \partial \sum_{t,s=1,\; t \ne s}^N p_{ts} \log \frac{p_{ts}} {q_{ts}}    } {\partial y_i}\\
&= -  \frac { \partial \sum_{t,s=1,\; t \ne s}^N p_{ts} \log q_{ts}  } {\partial y_i}\\
&= -  \frac { \partial \sum_{t,s=1,\; t \ne s}^N p_{ts} \log  \frac{ \tilde q_{ts}  } {Z_y} } {\partial y_i}\\
&= -  \frac { \partial \sum_{t,s=1,\; t \ne s}^N p_{ts} \log  \frac{ \tilde q_{ts}  } {Z_y} } {\partial y_i}\\
&= -  \frac { \partial \sum_{t,s=1,\; t \ne s}^N p_{ts} \log  \tilde q_{ts}   } {\partial y_i} +  \frac { \partial \log Z_y}{\partial y_i}  \\
&= -  2 \frac { \partial \sum_{j=1}^N p_{ij} \log  \tilde q_{ij}   } {\partial y_i} + \frac 1 {Z_y} \frac { \partial Z_y }{\partial y_i}  \\
&= -  2  \sum_{j=1}^N p_{ij}  \frac { \partial  \log  \tilde q_{ij}   } {\partial y_i} +  \frac 1 {Z_y}  \sum_{t,s=1,\; t \ne s}^N \frac { \partial \tilde q_{ts} }{\partial y_i}  \\
&= -  2  \sum_{j=1}^N p_{ij}  \frac { \partial  \log  \tilde q_{ij}   } {\partial y_i} +  2\frac 1 {Z_y}\sum_{j=1}^N \frac { \partial \tilde q_{ij} }{\partial y_i}  \\
&= -  2  \sum_{j=1}^N p_{ij}  \frac { \partial  \log  \tilde q_{ij}   } {\partial y_i} +  2\frac 1 {Z_y} \sum_{j=1}^N \tilde q_{ij}  \frac { \partial  \log  \tilde q_{ij}   } {\partial y_i}   \\
&= -  2  \sum_{j=1}^N p_{ij}  \frac { \partial  \log  \tilde q_{ij}   } {\partial y_i} +  2\sum_{j=1}^N q_{ij}  \frac { \partial  \log  \tilde q_{ij}   } {\partial y_i}   \\
&= -  2  \sum_{j=1}^N ( p_{ij} - q_{ij})  \frac { \partial  \log  \tilde q_{ij}   } {\partial y_i} 
\end{array}
\end{equation} 

\begin{equation}
\begin{array}{ll}
 \frac { \partial  \log  \tilde q_{ij}   } {\partial y_i}  &= \frac { \partial  \log \frac 1 { 1+ ||y_i - y_j||^2}  } {\partial y_i} \\
  &= - 2 \frac 1 { 1+ ||y_i - y_j||^2} (y_i - y_j)\\
  & = -2\tilde q_{ij} (y_i - y_j)
\end{array}
\end{equation}

\begin{equation}
\label{eq:grad-1}
\begin{array}{ll}
\frac { \partial L_{micro} } {\partial y_i}  &=  -  2  \sum_{j=1}^N ( p_{ij} - q_{ij})  \frac { \partial  \log  \tilde q_{ij}   } {\partial y_i}  \\
	&= 4\sum_{j=1}^N ( p_{ij} - q_{ij}) \tilde q_{ij} (y_i - y_j)
\end{array}
\end{equation} 

The second part is 

\begin{equation}
\begin{array}{ll}
\frac { \partial L_{macro} } {\partial y_i}  &= \frac { \partial KL(P_{macro}||Q_{macro})  } {\partial y_i}\\
&= -\frac { \partial \sum_{k,l=1,\; k \ne l}^K p_{macro, kl} \log{q_{macro, kl}}   } {\partial y_i}\\
&= -\frac { \partial \sum_{k,l=1,\; k \ne l}^K p_{macro, kl} \log{ \frac { \tilde q_{macro, kl} } {Z_c}   } }{\partial y_i}\\
&= -\frac { \partial \sum_{k,l=1,\; k \ne l}^K p_{macro, kl} \log{\tilde q_{macro, kl}   } }{\partial y_i}   + \frac { \partial \log Z_c} {\partial y_i} \\
&= -\frac { \partial \sum_{k,l=1,\; k \ne l}^K p_{macro, kl} \log{\tilde q_{macro, kl}   } }{\partial y_i}   + \frac 1 {Z_c} \frac { \partial Z_c} {\partial y_i} \\
&= -\frac { \partial \sum_{k,l=1,\; k \ne l}^K p_{macro, kl} \log{\tilde q_{macro, kl}   } }{\partial y_i}   + \frac 1 {Z_c} \frac { \partial \sum_{k,l=1,\; k \ne l}^K \tilde q_{macro, kl}  } {\partial y_i} \\
&= -\sum_{k,l=1,\; k \ne l}^K p_{macro, kl} \frac { \partial  \log{\tilde q_{macro, kl}   } }{\partial y_i}   + \frac 1 {Z_c}\sum_{k,l=1,\; k \ne l}^K  \frac { \partial  \tilde q_{macro, kl}  } {\partial y_i} \\
&= -\sum_{k,l=1,\; k \ne l}^K p_{macro, kl} \frac { \partial  \log{\tilde q_{macro, kl}   } }{\partial y_i}   + \frac 1 {Z_c}\sum_{k,l=1,\; k \ne l}^K \tilde q_{macro, kl}  \frac { \partial  \log \tilde q_{macro, kl}  } {\partial y_i} \\
&= -\sum_{k,l=1,\; k \ne l}^K p_{macro, kl} \frac { \partial  \log{\tilde q_{macro, kl}   } }{\partial y_i}   + \sum_{k,l=1,\; k \ne l}^K q_{macro, kl}  \frac { \partial  \log \tilde q_{macro, kl}  } {\partial y_i} \\
&= -\sum_{k,l=1,\; k \ne l}^K ( p_{macro, kl} - q_{macro, kl}) \frac { \partial  \log{\tilde q_{macro, kl}   } }{\partial y_i} 
\end{array}
\end{equation} 

\begin{equation}
\begin{array}{ll}
 \frac { \partial  \log{\tilde q_{macro, kl}   } }{\partial y_i}   &= \frac { \partial  \log \frac 1 { 1+ ||c_k - c_l||^2}  } {\partial y_i} \\
  &= - 2 \frac 1 { 1+ ||c_k - c_l||^2}(R_{ki} - R_{li})(c_k - c_l)\\
  & = -2\tilde q_{macro, kl} (R_{ki} - R_{li})(c_k - c_l)
\end{array}
\end{equation} 

\begin{equation}
\label{eq:grad-2}
\begin{array}{ll}
\frac { \partial L_{macro} } {\partial y_i}  &=   -\sum_{k,l=1,\; k \ne l}^K ( p_{macro, kl} - q_{macro, kl}) \frac { \partial  \log{\tilde q_{macro, kl}   } }{\partial y_i}  \\
    & = 2 \sum_{k,l=1,\; k \ne l}^K   ( p_{macro, kl} - q_{macro, kl})  \tilde q_{macro, kl} (R_{ki} - R_{li})(c_k - c_l)
\end{array}
\end{equation} 

The third part is 

\begin{equation}
\label{eq:grad-3}
\begin{array}{ll}
\frac { \partial L_{k-means} } {\partial y_i}  &=    \frac 1 N \frac { \sum_{k,j} \partial  R_{kj}||c_k - y_j||^2  } {\partial y_i}  \\
	&= 2 \frac 1 N \sum_k R_{ki}(y_i - c_k)
\end{array}
\end{equation} 

Substitute three parts (\ref{eq:grad-1}),  (\ref{eq:grad-2}),  (\ref{eq:grad-3}) into equation (\ref{eq:grad-total}), we get the gradient of GTSNE,

\begin{equation}
\begin{array}{ll}
\frac { \partial L(Y) } {\partial y_i} &= 4\sum_{j=1}^N ( p_{ij} - q_{ij}) \tilde q_{ij} (y_i - y_j) \\
&\quad + 2\alpha \sum_{k,l=1,\; k \ne l}^K   ( p_{macro, kl} - q_{macro, kl})  \tilde q_{macro, kl} (R_{ki} - R_{li})(c_k - c_l)  \\
&\quad +  2\beta   \frac 1 N\sum_{k = 1 }^K R_{ki}(y_i - c_k)
\end{array}
\end{equation} 
The derivation is finished.

\bibliographystyle{apalike}  
\bibliography{gtsne}  


\end{document}